\documentclass[letterpaper]{article}
\usepackage{preprintlayout}
\usepackage[hyphens]{url}
\usepackage{graphicx}
\usepackage{natbib}
\usepackage{caption}
\usepackage{amsmath}
\usepackage{booktabs}
\usepackage{multirow}

\title{Rethinking Class Imbalance for Single-Cell Foundation Models: A Systematic Benchmark Across Architectures and Long-Tail Loss Functions}
\author{
    Zeyu Dong \quad Jiahui Zhong
}
\preprintaffiliations{}

\begin{document}

\maketitle

\begin{abstract}
Single-cell foundation models (scGPT, scBERT, Geneformer) achieve cell-type
classification accuracy up to 97.5\% in our experiments, yet this
aggregate accuracy can mask systematic failure on rare, often disease-relevant cell
populations that long-tail loss functions are widely assumed to address. We
present a systematic benchmark of six long-tail loss functions
(cross-entropy, weighted CE, class-balanced loss, focal loss, LDAM,
logit-adjusted softmax) across three architectures and three datasets
(Multiple Sclerosis, Zheng68K, human Pancreas), totaling 162 controlled
training runs (3 backbones $\times$ 3 datasets $\times$ 6 losses $\times$ 3
seeds). The gap between overall accuracy, Macro-F1, and
rare-class recall under
plain cross-entropy is consistent across all nine
(architecture, dataset) settings, driven by dataset structure rather than
pretraining. Rare-class failure itself splits into two regimes with distinct
embedding-geometry signatures, visible before any loss is chosen: some
classes are recoverable by the right loss, while others retain linear
separability yet are absorbed into unrelated classes' neighborhoods
under every evaluated loss and architecture. Among the
recoverable classes, the efficacy of
reweighting is predicted by a class's \emph{absolute}
training-set size, rather than its share of the dataset or the
dataset's overall imbalance ratio. Class-balanced loss and LDAM are the most consistent choices across all
nine settings, while logit adjustment trades rare-class precision for
recall rather than improving both. Our results give both a reusable benchmark
and mechanism-grounded practical guidelines for combining foundation
models with imbalanced biological data.
\end{abstract}

\section{Introduction}

Foundation models for single-cell RNA sequencing (scRNA-seq), including scGPT
\citep{cui2024scgpt}, scBERT \citep{yang2022scbert}, Geneformer
\citep{theodoris2023geneformer}, scFoundation \citep{hao2024scfoundation}, and
CellPLM \citep{wen2024cellplm}, have become the default backbone for
cell-type annotation \citep{zhang2025scfmsurvey}. Reported accuracy
on standard benchmarks is often above 90\%, but this aggregate metric is
dominated by common cell types \citep{li2025annotationreview}. Rare cell populations,
which frequently carry critical disease-relevant signals, such as early-transitional or pathological
states \citep{zheng2024scnovel}, are systematically under-served, a pattern
first documented at the benchmark level by
\citet{alsabbagh2023foundation,wu2025biology}. \citet{alsabbagh2023foundation}
attempt to address it at the sampling-strategy level, and
\citet{naziri2025bias} document it in the specific context of
disease-state subgroup failure.

While long-tail failures in single-cell foundation models are increasingly
recognized, loss-level interventions have not been systematically investigated
across architectures. We benchmark six general-purpose long-tail strategies
\citep{lin2017focal,cui2019classbalanced,cao2019ldam,menon2021logitadjust}
to determine which are most consistently effective for rare cell populations,
whether benefits are backbone-dependent, and how embedding geometry and
training-set scarcity delimit the gains attainable through loss-level optimization.

Our contributions are fourfold. (1) We introduce the first systematic,
loss-function-level benchmark for class imbalance in single-cell
foundation models: a 162-run controlled study spanning a full
loss-function grid across three foundation-model architectures and
three general-purpose datasets, addressing both questions above. (2) We show that loss function
optimization alone is insufficient to recover certain rare classes, a
limitation explained via neural collapse theory
\citep{papyan2020prevalence} in Section~\ref{sec:diagnose}. (3) We find
that, among recoverable classes, absolute training-set size, rather
than relative frequency, predicts reweighting benefit, consistent with
findings in the general long-tail learning literature (the
effective-number-of-samples framework \citep{cui2019classbalanced},
computer-vision studies under imbalance
\citep{buda2018systematic,saini2023review}, and imbalance's multiple
characterizing axes \citep{gao2025survey}); we show this holds for
single-cell foundation models specifically, predicting how much a class
gains from replacing cross-entropy among classes with non-trivial
reweighting-based F1. (4) We show that different long-tail losses make
different, quantifiable trade-offs (recall for precision) that a
single aggregate metric such as Macro-F1 obscures, and translate this
into actionable guidelines for practitioners.

\section{Related Work}
\label{sec:relwork}

\citet{alsabbagh2023foundation} benchmark scGPT, scBERT, and Geneformer on
imbalanced cell-type annotation and find random oversampling the most
consistently effective mitigation, unlike undersampling or
imputation-based resampling. They attribute
Geneformer's weaker performance to its rank-based (as opposed to
expression-value-based) tokenization. Their study focuses on sampling
strategies across two datasets; we complement it with a loss-level
axis across three datasets, holding sample distribution and batch
composition fixed throughout to isolate the effect of the
loss-function axis. Other rare-cell annotation methods use synthetic
oversampling \citep{bej2021synthetic} or adaptive weight sampling
\citep{cheng2023scbalance}; unlike ours, these interventions jointly alter
the training distribution and classifier design.
A separate evaluation of four foundation models on an acute myeloid
leukemia dataset shows that strong aggregate F1 masks substantial
subgroup-level performance gaps (e.g., scGPT's F1 falling from above 0.90 on
healthy samples to below 0.75 on relapse cases), attributed to
pretraining-corpus bias toward healthy cells and calling for better
representation of disease-specific cell types in future pretraining data
\citep{naziri2025bias}. A complementary imbalance axis, demographic
composition rather than cell-type frequency, also measurably affects scGPT
fine-tuning outcomes \citep{alamin2025gender}. General-purpose
single-cell transcriptomics benchmarks cover similar
ground but do not focus on class-imbalance failure modes specifically
\citep{qi2025benchmark,qiu2025biollm}. \citet{liu2026scevalfm} compare ten single-cell foundation models across
eight downstream tasks but evaluate cell-type annotation only in
aggregate, without a rare-class or loss-function axis.

Outside single-cell biology, long-tail learning is a mature area
\citep{gao2025survey,dealvis2024survey,zhang2023survey}, spanning sampling-based approaches such as
SMOTE's synthetic minority oversampling \citep{chawla2002smote} and the
loss-level reweighting family we focus
on here; imbalance-aware benchmarking has also spread to other
structured-data domains, e.g., tabular data \citep{liu2025climb}.
Loss-level approaches re-derive the training objective to counteract
class frequency: inverse-frequency reweighting and its
effective-number-of-samples refinement, class-balanced loss
\citep{cui2019classbalanced}, focal loss's down-weighting of easy
examples \citep{lin2017focal}, LDAM's label-distribution-aware margin
\citep{cao2019ldam}, and logit adjustment's additive class-prior
correction \citep{menon2021logitadjust}. Within single-cell biology,
\citet{zhao2025celler} propose a dedicated loss function paired with a
purpose-built pretrained model and a new 40-million-cell dataset for
long-tailed annotation. Other work addresses long-tailed open-world annotation
through distribution-independent contrastive learning \citep{zhai2024distribution}.
We systematically benchmark six established
general-purpose losses across three existing pretrained backbones,
isolating how loss choice alone interacts with an otherwise unmodified
pretraining stage.
Representation-level approaches instead decouple feature learning from
classifier calibration \citep{kang2020decoupling}; \citet{weerasekara2026cellrefine}
apply this to single-cell foundation models via marker-gene priors
reshaping the embedding manifold post-pretraining, and other
methods build on this decoupling with
multiple distribution-aware experts \citep{wang2021ride}, supervised
contrastive learning \citep{cui2021paco}, or reflective learning
\citep{zhao2024ltrl}, combining representation-
and loss-level ideas in one pipeline. These directions require
either a second training stage on a frozen backbone or multiple
parallel classification heads. Our study focuses on loss-level
interventions that compose directly with pre-trained backbones,
offering a complementary perspective to architectural modifications.
A separate line of work
addresses long-tailed single-cell annotation through a class-incremental
lens, using generative replay to retain rare-class knowledge as new
cell types are introduced over time \citep{li2026scltcia}; we instead
focus on the fixed-class-set setting standard in cell-type annotation
benchmarks.

\section{Experimental Setup}
\label{sec:setup}

\paragraph{Backbones.} scGPT \citep{cui2024scgpt}, scBERT
\citep{yang2022scbert}, and Geneformer
\citep{theodoris2023geneformer}, each
fine-tuned with a linear classification head on top of the pretrained
encoder, using its default cell-level
aggregation: scGPT extracts a dedicated CLS token; Geneformer uses
HuggingFace's standard pooler (a transformed first-position hidden
state); scBERT instead passes the full per-gene-token sequence through
a convolution-and-fully-connected head, with no separate pooling step.
scGPT and Geneformer use full-parameter fine-tuning, with every
encoder and head parameter trainable. scBERT follows its default
protocol: the pretrained encoder is frozen except for its final
normalization layer and penultimate Performer block, alongside the
newly initialized classification head; the remainder of the encoder
is frozen. Each backbone thus uses its recommended fine-tuning
protocol rather than a single protocol imposed uniformly across
architectures. All three use a single
learning rate across every trainable
parameter, with no layer-wise decay. scGPT and Geneformer train for a fixed 15 epochs. scBERT trains
for up to 15 epochs with early stopping (patience 10 on validation
Macro-F1, matching scBERT's protocol). All three follow \citet{alsabbagh2023foundation}'s per-backbone
optimizer choice: Adam \citep{kingma2015adam} (lr $=10^{-4}$) for scGPT and
scBERT, AdamW \citep{loshchilov2019adamw} (lr $=5{\times}10^{-5}$, weight
decay $10^{-3}$) for Geneformer. Batch size follows \citet{alsabbagh2023foundation}'s per-backbone
protocol (32/1/12 for scGPT/scBERT/Geneformer respectively) on MS and
Pancreas; for scBERT on Zheng68K, we use gradient accumulation for an
effective batch size of 60, matching the batch-size-1 protocol applied
to MS and Pancreas. Every loss under test computes its class-level
term from fixed, globally precomputed training-set statistics, and
scBERT's Performer backbone normalizes per sample via LayerNorm;
neither depends on batch composition. All experiments were conducted on
NVIDIA RTX 4090 GPUs using bfloat16 mixed precision.

\paragraph{Datasets.} Multiple Sclerosis (MS, 18 cell types, top-3000
highly variable genes, underlying atlas from
\citealt{schirmer2019neuronal}) and Zheng68K (11 cell types, top-3000
highly variable genes, underlying atlas from \citealt{zheng2017massively})
follow the protocol of \citet{alsabbagh2023foundation}, which applies
this HVG reduction to MS but not Zheng68K; we extend it to Zheng68K as
well for cross-dataset consistency, selecting the panel from the
training split only and applying it unchanged to the test split. MS's
and Pancreas's gene panels are inherited pre-selected from public
tutorials. We retain two ambiguous MS labels as defined in the
source atlas. Human Pancreas (14 cell types, top-3000 highly variable
genes matching the other two datasets, underlying atlas from
\citealt{baron2016pancreas}) was introduced in this benchmark,
using scGPT's public tutorial annotation split, which leaves
three cell types (macrophage, Schwann, T cell) with zero test-set cells; we
exclude these from all evaluation metrics. All three datasets' gene panels are selected via
scanpy's \citep{wolf2018scanpy} \texttt{highly\_variable\_genes} (\texttt{cell\_ranger} flavor,
a variance-based feature-selection step) on library-size-normalized ($10^4$ counts per cell) and
$\ln(1+x)$-transformed expression, the same gene panel shared by
all three backbones per dataset, before any backbone-specific
tokenization. Geneformer's rank-based tokenizer requires values
proportional to raw counts rather than log-transformed ones: for
Zheng68K we used independently sourced raw counts directly; for MS and
Pancreas, where independent raw counts were unavailable, we recovered
count-proportional values via $\exp(x)-1$, exactly preserving gene
rankings since library-size normalization and this recovery both apply
the same per-cell scalar. Our three datasets' train/test
splits also differ by design: Zheng68K is class-proportion-preserving,
MS holds out entire patients (9 train/12 test donors, providing a
clinically realistic test of out-of-distribution generalization to
unseen individuals), and Pancreas follows its source tutorial's split.

\paragraph{Losses.} Cross-entropy (CE), inverse-frequency weighted CE,
class-balanced loss \citep{cui2019classbalanced} ($\beta=0.9999$), focal loss \citep{lin2017focal}
($\gamma=2.0$), LDAM
\citep{cao2019ldam} (maximum margin $0.5$, logit scale $30$, at
default), and logit-adjusted softmax
\citep{menon2021logitadjust}. Every loss uses default hyperparameters,
to evaluate each loss family's
off-the-shelf robustness on this benchmark.

\paragraph{Protocol.} For each of the $3 \times 3 \times 6 = 54$ (backbone,
dataset, loss) cells we train 3 seeds and report the mean. For each run, we
select the checkpoint with the best Macro-F1 on a held-out validation
set (10\% of the training data)
and evaluate it once on the test set. Class predictions are assigned by
argmax over each model's output logits. A class is
considered rare if its training-split frequency is below 5\%
per domain experience; let
$\mathcal{R}\subseteq\{1,\dots,C\}$ denote the resulting set of rare
classes in a given dataset. We report accuracy, Macro-F1,
\begin{equation}
\text{Macro-}F_1 = \frac{1}{C}\sum_{c=1}^{C} F_{1,c},
\label{eq:macrof1}
\end{equation}
the unweighted mean of each class's F1 score across all $C$ evaluated
classes, and rare-class recall,
\begin{equation}
\text{Recall}_{\mathcal{R}} = \frac{1}{|\mathcal{R}|}\sum_{c\in\mathcal{R}} \text{Recall}_c,
\label{eq:rarerecall}
\end{equation}
the macro-averaged recall restricted to $\mathcal{R}$. Rare-class precision
is defined analogously, using precision$_c$ in place of recall$_c$. Rare-class
AUPRC (Table~\ref{tab:main}) instead macro-averages each class's
one-vs-rest area under the precision-recall curve over $\mathcal{R}$.

\section{Systematic Evaluation}
\label{sec:results}

\subsection{Aggregate Accuracy Overstates Rare-Class Performance, Regardless of Architecture}
\label{sec:accgap}

\begin{figure}[t]
\centering
\includegraphics[width=\columnwidth]{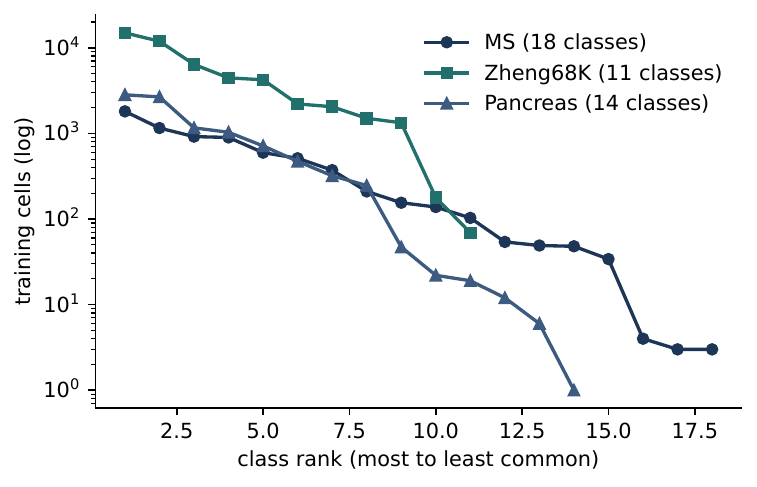}
\caption{Training-set class rank vs.\ count for all three datasets (log
scale). MS's and Pancreas's smallest classes reach a minimum of
single-digit sample counts, while Zheng68K's smallest class still has
$\sim$70 training cells.}
\label{fig:classdist}
\end{figure}

\begin{figure*}[t]
\centering
\makebox[\textwidth][c]{\includegraphics[width=0.43\textwidth]{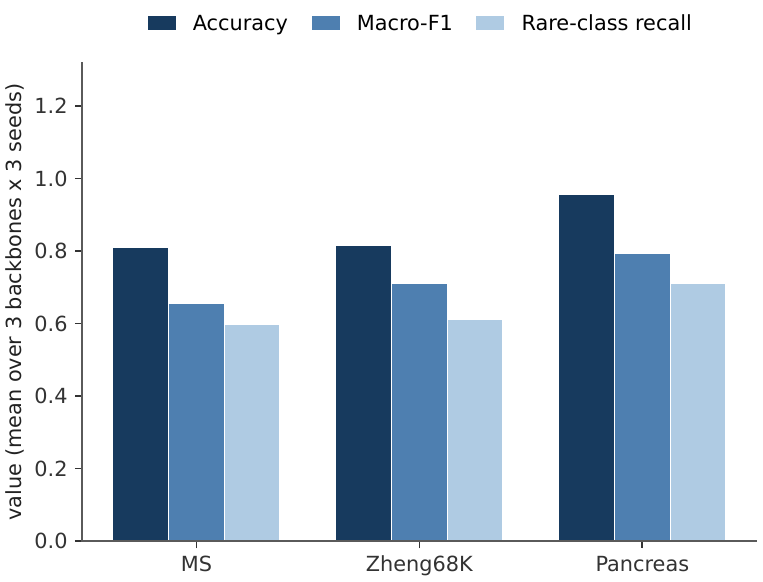}\hspace{0.04\textwidth}\includegraphics[width=0.43\textwidth]{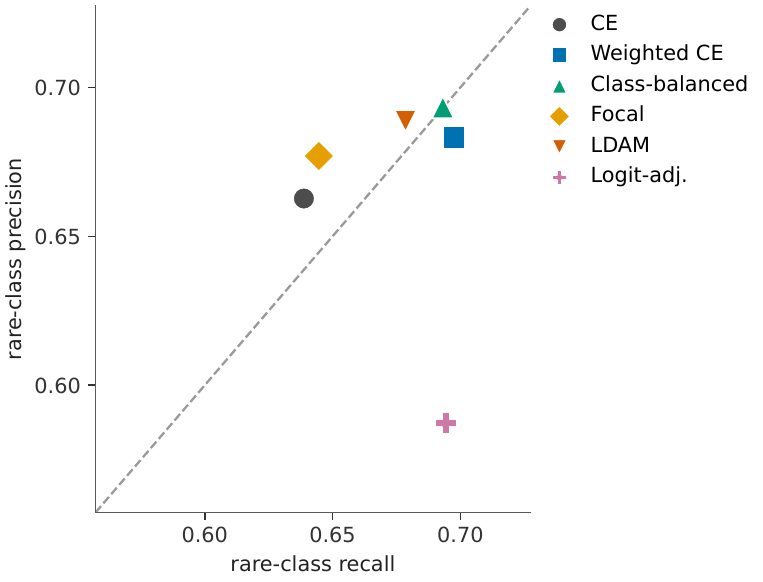}}
\caption{\textbf{Left:} per-dataset breakdown under plain CE (mean of 9
backbone$\times$seed runs per bar); Pancreas has the highest accuracy but
also the largest accuracy-to-rare-recall drop. \textbf{Right:}
rare-class precision vs.\ recall by loss (Section~\ref{sec:precision}); dashed line is
$y=x$. Points above trade recall for precision, points below trade
precision for recall.}
\label{fig:gap}
\end{figure*}

The raw training-set class distribution for all three datasets
(Figure~\ref{fig:classdist}) shows Pancreas is the most extreme by relative
measures (imbalance ratio 2822:1 and 64\% of its classes below the 5\%
rare-class threshold, both higher than MS's 603:1 and 61\%); however,
MS exhibits the greatest benefit from reweighting. Figure~\ref{fig:gap} (left)
shows accuracy, Macro-F1, and rare-class recall under plain cross-entropy,
averaged over all three backbones, for each dataset. Across all nine
(backbone, dataset) combinations, accuracy exceeds Macro-F1, which in
turn exceeds rare-class recall. This gap is present regardless of which
foundation model is used, indicating it is an inherent property of
the data's imbalance structure. (Its plain-CE Macro-F1 for scBERT on
MS is 0.639, differing from 0.858 reported by
\citet{alsabbagh2023foundation}; this variance arises from adoption of
scBERT's default configuration across Gene2Vec
\citep{du2019gene2vec}, checkpoint-loading, batch size, and optimizer
schedule.)

\subsection{Loss-Function Comparison}
\label{sec:losscomp}

Table~\ref{tab:main} reports Macro-F1 and rare-class recall for all six
losses, averaged across backbones and seeds. Loss rankings below are still computed
within each (backbone, dataset) cell before averaging, so this
cross-backbone variance does not confound them. Ranking each loss by
its average position across all nine (backbone, dataset) cells, class-balanced loss
(mean rank 2.4 of 6) and LDAM (2.6) are the most consistent generalists,
ahead of the remaining four losses; neither is the top performer on
every dataset column (weighted CE has the best MS Macro-F1 at 0.721, and
LDAM itself the best Pancreas Macro-F1 at 0.819), but both consistently rank
near the top across all nine cells rather than exhibiting severe
dataset-specific variance. Ranked instead by rare-class macro-AUPRC
(a threshold-free measure of ranking quality across the full
precision-recall curve that avoids committing to any single decision
threshold), class-balanced loss remains the strongest (0.734). LDAM,
in contrast, yields the lowest performance (0.690) despite its strong
F1-based mean rank, consistent with a margin-based objective that
shapes decision boundaries without directly calibrating class
probabilities. Logit adjustment gives
the largest rare-recall gain on Zheng68K (0.685 vs.\ the next-best 0.649)
and ties for second on Pancreas (0.766 vs.\ LDAM's 0.767; paired $t$-test
across the nine backbone-seed cells, $t=0.04$, $p=0.97$), trailing only
weighted CE's 0.772. It is
nonetheless the weakest generalist overall (mean rank 5.2), trading
broad consistency for rare-class-specific recall. The relative Macro-F1 improvement of the best
loss over CE is largest on MS (+9.9\%) and smallest on Zheng68K (+0.9\%),
even though Zheng68K is not the most imbalanced dataset in absolute-count
terms, as analyzed in Section~\ref{sec:abscount}.

\begin{table*}[t]
\centering
\footnotesize
\setlength{\tabcolsep}{4.8pt}
\begin{tabular*}{\textwidth}{@{\extracolsep{\fill}}lccccccccc}
\toprule
 & \multicolumn{2}{c}{MS} & \multicolumn{2}{c}{Zheng68K} & \multicolumn{2}{c}{Pancreas} & & & \\
\cmidrule(lr){2-3} \cmidrule(lr){4-5} \cmidrule(lr){6-7}
Loss & F1 & $\text{Recall}_{\mathcal{R}}$ & F1 & $\text{Recall}_{\mathcal{R}}$ & F1 & $\text{Recall}_{\mathcal{R}}$ & $\Delta$F1 & Rare AUPRC & Mean Rank \\
\midrule
Cross-entropy   & 0.656$_{\pm.054}$ & 0.596$_{\pm.072}$ & 0.711$_{\pm.026}$ & 0.610$_{\pm.054}$ & 0.794$_{\pm.052}$ & 0.710$_{\pm.106}$ & --      & 0.712$_{\pm.083}$ & 4.56 \\
Weighted CE     & \textbf{0.721}$_{\pm.035}$ & 0.687$_{\pm.042}$ & 0.704$_{\pm.041}$ & 0.634$_{\pm.050}$ & 0.816$_{\pm.039}$ & 0.772$_{\pm.065}$ & $+0.026$ & 0.728$_{\pm.066}$ & 2.78 \\
Class-balanced  & 0.716$_{\pm.042}$ & 0.685$_{\pm.047}$ & 0.708$_{\pm.042}$ & 0.649$_{\pm.050}$ & 0.816$_{\pm.028}$ & 0.746$_{\pm.068}$ & $+0.026$ & \textbf{0.734}$_{\pm.062}$ & \textbf{2.44} \\
Focal           & 0.662$_{\pm.059}$ & 0.605$_{\pm.071}$ & \textbf{0.718}$_{\pm.027}$ & 0.625$_{\pm.042}$ & 0.787$_{\pm.034}$ & 0.704$_{\pm.081}$ & $+0.002$ & 0.719$_{\pm.068}$ & 3.44 \\
LDAM            & 0.693$_{\pm.050}$ & 0.644$_{\pm.066}$ & 0.717$_{\pm.031}$ & 0.625$_{\pm.048}$ & \textbf{0.819}$_{\pm.021}$ & 0.767$_{\pm.047}$ & $+0.022$ & 0.690$_{\pm.095}$ & 2.56 \\
Logit-adjusted  & 0.650$_{\pm.045}$ & 0.632$_{\pm.061}$ & 0.704$_{\pm.038}$ & \textbf{0.685}$_{\pm.040}$ & 0.761$_{\pm.047}$ & 0.766$_{\pm.062}$ & $-0.016$ & 0.700$_{\pm.075}$ & 5.22 \\
\bottomrule
\end{tabular*}
\caption{Macro-F1 and rare-class recall ($n{=}9$: 3 backbones $\times$ 3
seeds; $\pm$ is the standard deviation across these nine cells) per loss
and dataset, plus mean $\Delta$F1 over CE
(Eq.~\ref{eq:deltaf1}), rare-class macro-AUPRC (area under the
precision-recall curve per rare class, macro-averaged, mean over the
same nine cells; $n{=}24$ for cross-entropy, $n{=}27$ for every other
loss), and mean rank (1=best, 6=worst) across all nine cells.
Bold marks the best value per column.}
\label{tab:main}
\end{table*}

\subsection{Reweighting Benefit Is Predicted by Absolute Rare-Class Sample Count}
\label{sec:abscount}

\begin{figure*}[t]
\centering
\mbox{\hspace*{0.07\textwidth}\includegraphics[width=0.36\textwidth]{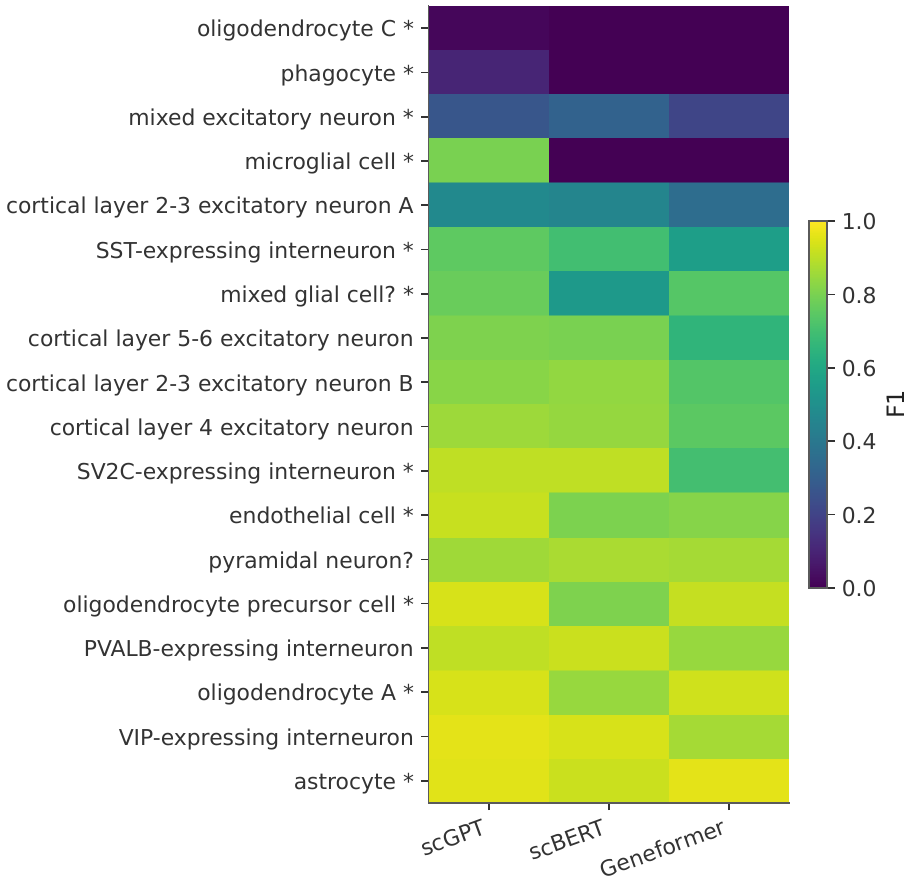}\hspace{0.05\textwidth}\includegraphics[width=0.41\textwidth]{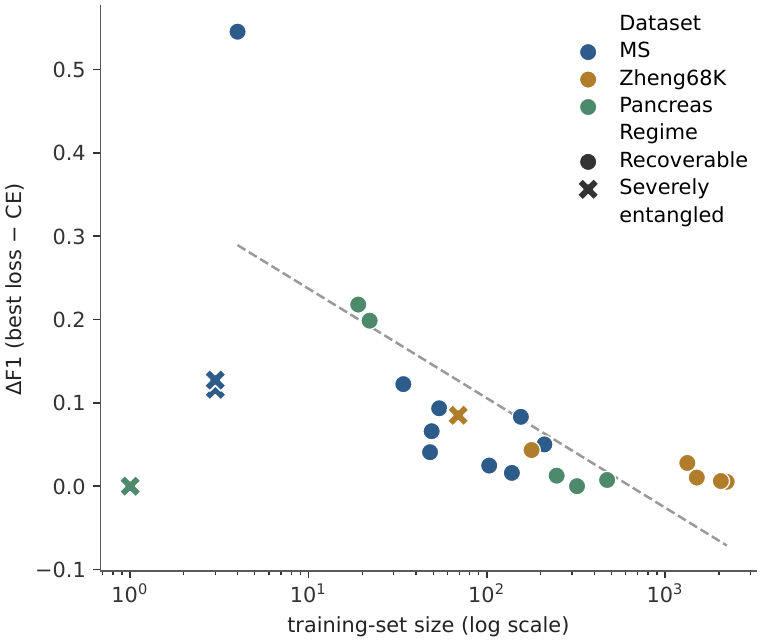}\hspace*{0.07\textwidth}}
\caption{\textbf{Left:} per-class F1 heatmap (CE, MS), rows sorted by
mean F1 across the three backbones, hardest (top) to easiest (bottom);
a class with high F1 on one backbone but low F1 on the other two can
still rank as hard. \texttt{*} marks a rare class
(Section~\ref{sec:setup}). The two hardest classes are dark across all three columns
since no backbone architecture recovers them; classes further down show
lighter, more backbone-dependent shading (three-way split,
Section~\ref{sec:diagnose}). \textbf{Right:} per-class $\Delta$F1 against training-set
size (log $x$-axis) for all rare classes across all three datasets; color
marks dataset. Circles (fit line, $n{=}19$, $r{=}{-}0.77$) are the recoverable regime;
$\times$ markers show the four severely entangled classes (two
from MS, one each from Zheng68K and Pancreas); the same near-floor
regime recurs across all three datasets (Section~\ref{sec:abscount}).}
\label{fig:count}
\end{figure*}

To determine whether the best loss's Macro-F1 gain over CE
(Section~\ref{sec:losscomp}: $+9.9\%$ on MS, $+0.9\%$ on Zheng68K) is
spread evenly across rare classes or concentrated in a few, we
evaluate what predicts each rare class's gain from switching away
from cross-entropy. For each rare
class $c$, we compute
\begin{equation}
\Delta F_{1,c} = \max_{\ell \in \mathcal{L}} F_{1,c}^{(\ell)} - F_{1,c}^{(\text{CE})},
\label{eq:deltaf1}
\end{equation}
where $\mathcal{L}$ is the set of six losses under test and $F_{1,c}^{(\ell)}$
is class $c$'s F1 under loss $\ell$, averaged over backbones and seeds. We
then correlate $\Delta F_{1,c}$ with that class's absolute training-set size. We
exclude rare classes whose best-available-loss F1 is itself near zero
(mean $<0.2$ across backbones, $n=4$ classes) from this regression, since
these represent the severely entangled regime
(Section~\ref{sec:diagnose}). Figure~\ref{fig:count} (right) shows a strong,
significant negative Pearson correlation ($r=-0.77$, $p<0.001$, $n=19$ rare
classes) between each rare class's log-transformed absolute training-set
size and its reweighting gain. Reintroducing these four severely entangled classes yields a weaker
but still significant correlation ($r=-0.53$, $p=0.010$, $n=23$),
demonstrating the trend's robustness to this exclusion. This inverse relationship between absolute sample count and reweighting benefit directly explains the dataset-level pattern in
Section~\ref{sec:losscomp}: MS, whose rare classes have the fewest training examples in
absolute terms despite a comparable imbalance \emph{ratio} to Zheng68K,
exhibits the greatest benefit from reweighting. This relationship remains negative and
significant across a range of severely entangled exclusion thresholds
($r=-0.53$ to $-0.78$, $p<0.01$, $n=18$--$23$) and rare-class-frequency
cutoffs ($r=-0.72$ to $-0.80$, $p<0.001$, $n=14$--$30$).

\subsection{Logit Adjustment Trades Rare-Class Precision for Recall Gains}
\label{sec:precision}

Figure~\ref{fig:gap} (right) plots rare-class precision against rare-class
recall, averaged over all nine settings, for each loss. Class-balanced
loss shows the largest joint gain in both precision and recall relative
to CE (precision
$+0.031$, recall $+0.054$), followed closely by LDAM ($+0.026$,
$+0.040$), both improving precision and recall simultaneously. Logit
adjustment improves recall by a similar margin but incurs a substantial
precision penalty: it has
the lowest rare-class precision of all six losses. This indicates it
achieves its recall gains primarily by shifting the decision threshold
rather than fundamentally improving feature separability, leading to
more false
positives. Non-rare-class precision and recall are comparatively stable
across loss choice (1.1- and 2.1-point spreads respectively across all six
losses), versus a 10.6-point spread in rare-class precision and a 5.9-point
spread in rare-class recall, indicating the differences between losses
are concentrated in rare-class behavior.

\subsection{Computational Efficiency}
\label{sec:efficiency}

Loss-level reweighting introduces negligible computational overhead:
across all nine (backbone, dataset) settings and five reweighting-based
losses (45 comparisons), wall-clock training time stays within 3.2\% of plain
cross-entropy, and under 1\% in most cases. Backbone selection affects
cost far more than loss choice: on MS under plain CE (seed 0), scBERT
requires 35.5 minutes (batch size 1) versus 6.2 for scGPT and 1.9
for Geneformer, while scGPT's peak GPU memory (8.0 GB) is roughly $6\times$
Geneformer's (1.3 GB) and $19\times$ scBERT's (0.4 GB).

\section{Diagnostic Analysis}
\label{sec:diagnosis}
\label{sec:diagnose}

\subsection{Two Distinct Rare-Class Failure Regimes}
\label{sec:regimes}

Figure~\ref{fig:count} (left) shows per-class F1 for CE across the three
backbones on MS, sorted from hardest to easiest. Two classes exhibit
near-zero F1 scores irrespective of backbone or loss (oligodendrocyte C,
phagocyte). Both have
only 3 training examples, too few for any of the reweighting schemes we
test to compensate for. Several other rare classes (e.g., microglial cell) show F1
ranging from 0 to nearly 1 depending on loss and seed. For these specific
classes, the choice of loss function substantially affects performance,
whereas the remaining rare classes already perform well under plain CE
regardless of loss choice. This three-way split
(severely entangled, loss-sensitive, and loss-insensitive) recurs on Zheng68K and
Pancreas and suggests that reporting a single
aggregated rare-class metric can obscure where a given method is
effective. To test whether this failure is specific to
foundation-model pretraining or persists across representations,
we additionally evaluate two non-foundation-model
baselines: a distance-weighted 5-nearest-neighbor classifier
\citep{coverhart1967nearestneighbor,dudani1976distanceweighted} and a
class-balanced random forest \citep{breiman2001randomforests,chen2004randomforest}
with 300 trees, both fit directly on raw gene-expression
features with no intervening learned representation step (e.g., PCA or
a variational embedding). Both yield an F1 score of 0 on these same
classes, matching every loss function's fully trained classifier head;
the failure is therefore not specific to pretrained representations.
Both baselines are otherwise non-trivial on MS overall
(Macro-F1 0.28 for the 5-NN, 0.65 for the random forest, comparable to
the foundation models' plain-CE Macro-F1), confirming this is not
merely a degenerate-baseline artifact.

\subsection{Geometric Signatures and Neighborhood Absorption}
\label{sec:geometry}

\begin{figure*}[t]
\centering
\includegraphics[width=0.95\textwidth]{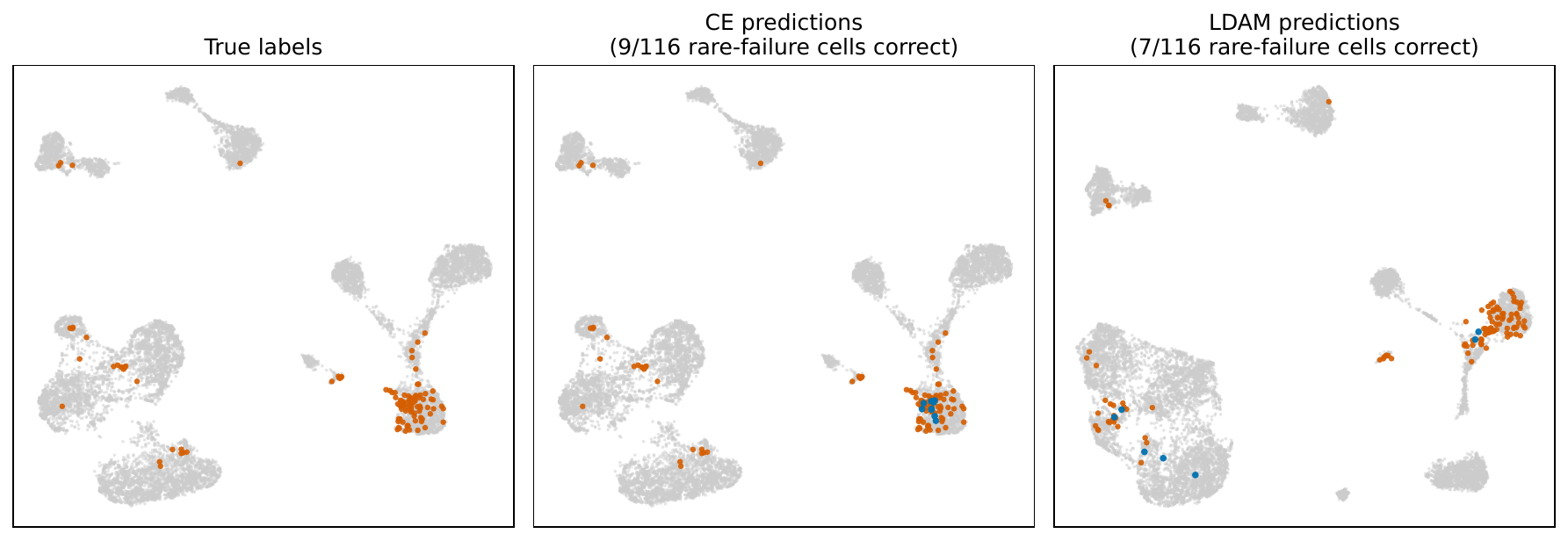}
\caption{UMAP \citep{healy2024umap} of MS test-set embeddings (scGPT), with the two
severely entangled classes' 116 test-set cells (oligodendrocyte
C: 34, phagocyte: 82) highlighted against all other classes in gray
throughout all three panels.
\textbf{Left:} true labels in CE's embedding space; these 116 cells
appear in orange. \textbf{Center:} the same 116
cells in the same CE embedding space, now colored by CE's prediction
correctness (blue = correctly classified, orange = misclassified).
\textbf{Right:} the same 116 cells in LDAM's own, separately fine-tuned
embedding space, colored by LDAM's prediction correctness, since CE
and LDAM are separately fine-tuned checkpoints. Both losses recover only
a small minority of these cells (9/116 under CE, 7/116 under LDAM, both
blue), confirming the failure is rooted in the embedding geometry
itself, not a specific classifier head.}
\label{fig:umap}
\end{figure*}

We analyze the embedding geometry
and confusion patterns of the two severely entangled MS classes to
understand their uniform failure across CE, LDAM, and logit-adjusted
softmax: they are
absorbed into unrelated classes' regions of representation space rather than
forming their recognizable cluster, consistent with too few training
examples to learn a distinguishable representation, independent of
backbone or loss (Figure~\ref{fig:umap}). Extending this analysis to all three backbones under CE,
both classes are consistently misclassified
into the \emph{same} pair of neighboring classes regardless of architecture.
Oligodendrocyte A and an unspecified mixed glial type jointly account for
91--97\% of all phagocyte misclassifications across scGPT, scBERT, and
Geneformer (individually 44--76\% and 21--47\% respectively, varying in
relative split while the same two classes dominate throughout).
Oligodendrocyte C is absorbed into excitatory-neuron classes for
75--89\% of its errors in all three. This cross-architecture agreement is consistent with
real transcriptional overlap between these cell types: phagocyte is
identified by ingested myelin/oligodendrocyte RNA
\citep{schirmer2019neuronal}, rather than an
architecture-specific failure mode. The same pattern appears at the representation
level: phagocyte's embedding centroid is the
nearest of all 17 other MS classes to \emph{both} of its confusion
partners in every one of six (backbone, loss) combinations we evaluated
(scGPT, scBERT, Geneformer $\times$ CE, LDAM), indicating a topology-driven
failure rather than a classifier decision-boundary failure.

Phagocyte and oligodendrocyte C differ in their geometric failure
modes. Specifically, 72.5\% of phagocyte cells' nearest non-self-class
neighbors belong to oligodendrocyte A alone (3 unique neighbor classes,
normalized entropy 0.71), a concentrated absorption consistent with
data scarcity. Conversely, oligodendrocyte C's nearest neighbors spread
across 10 classes (normalized entropy 0.90), extending beyond mere
data scarcity. Biologically, oligodendrocyte C is an MS-specific pathological
subcluster; its confusion with excitatory neurons reflects shared
self-antigen-presentation (B2M, HLA-C) and cell-stress activation in MS
lesions \citep{schirmer2019neuronal}. This geometric-proximity pattern
generalizes beyond MS, though less uniformly: under scGPT/CE, Zheng68K's CD14+ Monocyte and
Dendritic cells are mutually each other's rank-1/rank-2 nearest
centroid ($p\approx0.022$), while on Pancreas the effect is present but
weaker, with MHC class II's confusion partner ranking only second nearest.
Geometric proximity consistently contributes to confusion across
datasets, although its magnitude varies.

We also quantify the within-class/between-class geometry using the standard
neural collapse trace ratio \citep{papyan2020prevalence}
\begin{equation}
\mathrm{NC1} = \frac{\mathrm{tr}(\Sigma_W)}{\mathrm{tr}(\Sigma_B)},
\label{eq:nc1}
\end{equation}
where $\Sigma_W$ is the
average within-class scatter and $\Sigma_B$ is the between-class scatter of
class centroids around their group mean (lower NC1 indicates tighter,
better-separated clusters; the ratio is the reciprocal of the
Fisher Discriminant Ratio from classical pattern recognition). Following
prior work characterizing this geometry directly under class imbalance
\citep{fang2021minoritycollapse,zhang2024allnc}, we adopt this trace
ratio to geometrically characterize frozen test-set embeddings.
Restricted to MS's 11 rare classes under plain
CE, oligodendrocyte C shows substantially higher NC1 than most other rare
classes on every backbone (scGPT: 1.19, scBERT: 1.30, Geneformer: 1.79;
second-highest of the 11 rare classes on average, behind only mixed
excitatory neuron), quantitatively confirming it lacks the tight,
well-separated clustering observed in most other rare classes,
consistent with the confusion and nearest-neighbor evidence above.
Phagocyte's NC1, by contrast, is not elevated (0.17--0.32, ranking 8th
of 11): its small cell count means a handful of tightly clustered
cells embedded inside another class's region can produce low
within-class scatter despite total confusability, a signature the
nearest-neighbor evidence above captures instead. This contrast
replicates across all three MS backbones. Together, the
confusion-pattern, NC1, and nearest-neighbor evidence distinguish MS's
two severely entangled classes from its other rare classes, each
through a different geometric signature.

\subsection{Linear Probing Reveals Unused Discriminative Signal}
\label{sec:probe}

To investigate whether ``severely entangled'' reflects a genuine absence of
discriminative signal, or merely a failure of the six loss functions to
exploit signal that is present, we fit an independent linear probe
\citep{alain2017probes} for oligodendrocyte C, phagocyte, and (as a
loss-sensitive control) microglial cell on each backbone's
CE-fine-tuned embeddings: a class-balanced logistic regression evaluated by 5-fold
stratified cross-validation within the frozen test-set embeddings
(measuring CV-internal separability, not train/test generalization).

For the two entangled classes, the probe recovers
non-trivial F1 with high recall but low precision (oligodendrocyte C:
0.175--0.465, phagocyte: 0.248--0.462 across backbones), and in all six
(backbone, class) combinations this \emph{exceeds} the best F1 the
model's fully trained classifier head achieves under any of the six
losses (e.g., scGPT/oligodendrocyte C: probe 0.465 vs.\ best trained
0.329; Geneformer/oligodendrocyte C: probe 0.175 vs.\ 0.000). For the
loss-sensitive control class microglial cell, this margin shrinks
sharply and, for two of three backbones, reverses: scGPT's probe still
outperforms its trained classifier, but by only 4.3 points, versus 13.6
and 8.3 points for oligodendrocyte C and phagocyte respectively, while
scBERT's probe underperforms by 8.5 points (0.634 vs.\ 0.719) and
Geneformer's by 1.6 points (0.844 vs.\ 0.860).

Discriminative signal
therefore survives in the frozen representation for the two severely
entangled classes despite only 3 training examples each. The reweighting-based losses we evaluate
leave part of this signal unused, placing the bottleneck partly in
loss design rather than solely in data scarcity. Since the probe and
the model's classifier head are fit on the same fine-tuned
embeddings, this localizes the bottleneck to classifier training
rather than to a collapse of the representation itself, consistent
with the motivation behind decoupling classifier calibration from
representation learning \citep{kang2020decoupling}.

\section{Discussion and Limitations}

Given that the rare cell types most affected by this bottleneck are
disproportionately disease-relevant \citep{naziri2025bias,zheng2024scnovel,schilder2026phenome},
a foundation model that reports strong aggregate accuracy can still be
unreliable on the subpopulations most relevant to a clinical or discovery
pipeline. A severely entangled
class identified by the embedding-geometry check in
Section~\ref{sec:diagnose} signals a data-scarcity bottleneck better
addressed by collecting more data or abstaining (a reject-option
strategy) than by further loss engineering.

\paragraph{Practical Guidelines.} First, default to class-balanced loss when a rare class is
loss-sensitive: it yields the most consistent gains in both
classification quality and calibration (Section~\ref{sec:losscomp}); LDAM is a close
second by rank but shows the weakest calibration. Second, treat logit
adjustment as a trade-off mechanism for precision and recall rather
than a default choice: its
recall gains come at a substantial precision cost (Section~\ref{sec:precision}), so reserve it
for settings where missing a rare-class instance is costlier than a
false positive. Third, embedding geometry, such as the NC1 trace ratio
and nearest-neighbor confusion patterns computed before any loss is
chosen, predicts which rare classes will fail regardless of loss choice
(Section~\ref{sec:diagnose}); for those that do, a linear probe on the
frozen embeddings distinguishes unused signal, warranting further loss
engineering, from genuine data scarcity.

\paragraph{Biological Significance of the Loss-Sensitive Regime.} This
regime has direct biological implications: although microglia
mediate demyelination in MS \citep{zhang2023microglia}, detecting them
under CE remains architecture-dependent. They are
well recovered on scGPT (F1 0.73--0.84)
but undetected on scBERT and Geneformer (F1 $=$ 0 across seeds).
Class-balanced loss and weighted CE recover this population
consistently on Geneformer (F1 0.84--0.87) and partially on scBERT
(F1 0.86--0.92 in two seeds; 0.32--0.39 in one).

\paragraph{Limitations and Future Work.} We evaluate three pretrained architectures (scGPT,
scBERT, Geneformer); other existing single-cell foundation models,
such as scFoundation and CellPLM \citep{hao2024scfoundation,wen2024cellplm},
remain untested. While our linear-probe analysis
(Section~\ref{sec:diagnose}) demonstrates that discriminative signal is
linearly separable in the frozen representation, a more expressive,
nonlinear classifier head might extract
additional signal beyond what the six losses we test currently
achieve. Finally, this study isolates cell-type class imbalance
specifically; compounding variables such as
demographic composition \citep{alamin2025gender} or batch effects
\citep{maan2024imbalance} require future investigation.

\section{Conclusion}

We present a 162-run benchmark of six long-tail losses across three
single-cell foundation model architectures and three datasets.
Rare-class failure
manifests as two diagnosable regimes, loss-sensitive and severely
entangled, the latter being discernible in embedding geometry and
unresolved by the evaluated losses. Among the recoverable
classes, a rare class's absolute training-set size, rather than
its share of the dataset, predicts its benefit from reweighting. Class-balanced loss
and LDAM are the most consistent choices across all nine settings,
whereas logit adjustment trades rare-class precision for recall rather
than improving both.

\paragraph{Code Availability.} Implementation and experiment configurations
are available at \url{https://github.com/jz890/sc-imbalance}.

\end{document}